\documentclass{article}

\usepackage{floatrow}
\usepackage{graphicx}
\usepackage{float}
\usepackage{biblatex}
\usepackage{csquotes}
\usepackage[utf8]{inputenc}
\usepackage{biblatex}
\usepackage[preprint,nonatbib]{neurips_2021}

\usepackage[utf8]{inputenc} 
\usepackage[T1]{fontenc}    
\usepackage{hyperref}       
\usepackage{url}            
\usepackage{booktabs}       
\usepackage{amsfonts}       
\usepackage{nicefrac}       
\usepackage{microtype}      
\usepackage{xcolor}         
\usepackage{amsmath}
\usepackage{subfigure}
\usepackage{caption}
\usepackage{sidecap}
\sidecaptionvpos{figure}{c} 

\title{GrowSpace: Learning How to Shape Plants}

\author{
  Yasmeen Hitti\\
  McGill University, Mila\\
  \texttt{yasmeen.hitti@mail.mcgill.ca} \\
  \And
  Ionelia Buzatu \\
  Johannes Kepler Universität Linz\\
  \texttt{buzatuionelia@gmail.com} \\
  \AND
  Manuel Del Verme\\
  McGill University, Mila \\
  \texttt{manuel.delverme@gmail.com} \\
  \And
  Mark Lefsrud \\
  McGill University \\
  \texttt{mark.lefsrud@mcgill.ca} \\
  \And
  Florian Golemo\\
  Université de Montréal, Mila, Element AI\\
  \texttt{fgolemo@gmail.com} \\
  \And
  Audrey Durand\\
  Université Laval, Mila\\
  \texttt{audrey.durand@ift.ulaval.ca} \\
}

\newcommand{\growspace}{GrowSpace} 

\begin{document}

\maketitle

\begin{abstract}
Plants are dynamic systems that are integral to our existence and survival. Plants face environment changes and adapt over time to their surrounding conditions. We argue that plant responses to an environmental stimulus are a good example of a real-world problem that can be approached within a reinforcement learning (RL) framework. With the objective of controlling a plant by moving the light source, we propose \growspace{}, as a new RL benchmark. The back-end of the simulator is implemented using the Space Colonisation Algorithm, a plant growing model based on competition for space. Compared to video game RL environments, this simulator addresses a real-world problem and serves as a test bed to visualize plant growth and movement in a faster way than physical experiments. \growspace{} is composed of a suite of challenges that tackle several problems such as control, multi-stagelearning, fairness and multi-objective learning. We provide agent baselines alongside case studies to demonstrate the difficulty of the proposed benchmark. 
\end{abstract}

\section{Introduction}
Advancements in Reinforcement Learning (RL)~\cite{sutton2018reinforcement} are in part from comparing algorithms on commonly used benchmarks such as the Atari Learning Environment~\cite{bellemare2013arcade}. Despite Atari offering a range of control tasks and fast running speeds, state-of-the-art algorithms such as MuZero\cite{schrittwieser2020mastering} and recently Muesli\cite{hessel2021muesli} have reached 400\% human-level performance. This raises doubts on the difficulty of popular benchmarks since they do not always translate to real-world applications and inherently fail to capture the generalization performance of RL algorithms for real-world deployment~\cite{henderson2018deep}. The RL community needs new challenges and simulation-driven benchmark environments that tackle practical problems or at least a clear analogy to the real-world and introduce higher stochasticity \cite{dulac2019challenges}.

Currently there are a limited number of benchmarks that represent real-world systems since they are hard to design and learning from the physical world is difficult~\cite{mahmood2018benchmarking,dulac2019challenges}. Their complexities stem from high operating costs, their slow movements, and their limited amount of data~\cite{dulac2020empirical}. Simulators have provided a proxy to real-world systems and have demonstrated success in optimization of control tasks in robotics~\cite{rusu2017sim}. 

We direct our interest on plants, which similarly to robots, need to interact with their environment. Plants are complex and sense their surroundings through actuation and sensing systems~\cite{fratzl2008cellulose}. As biological systems, they actuate their movement as a response to an external stimulus such as light~\cite{burgert2009actuation}. Their spatial reorientation and growth towards light is a tropic response because their movement is influenced by the direction of the light source~\cite{liscum2014phototropism}. Recently, the idea of controlling plant growth through light manipulation has been investigated for the development of bio-hybrid systems such as living structures~\cite{wahby2018autonomously}. The control of a biological agent, presents a set of interesting problems which translate well to the RL community, such as: continuous control~\cite{recht2019tour}, multi-stagelearning~\cite{yang2018hierarchical}, multi-objective learning~\cite{van2014multi}, and fairness in a multiple plant setting~\cite{jabbari2017fairness}.

In this work, we introduce \growspace{}, a new RL environment that enables the control of procedurally generated plant structures. 
This benchmark is based on real plant responses to light and leverages this response to address a set of diverse challenges that are beyond the scope of bio-engineering. 
We bring attention to a set of four different challenges that range from classic control to fairness. \growspace{} is an environment that spans across different fields such as plant science, agriculture, RL, and robotics. 

The primary contributions of this paper include: (i) \growspace{}\footnote{\url{https://github.com/YasmeenVH/growspace}}, an OpenAI Gym-compatible environment~\cite{brockman2016openai} for RL, plant science, and robotics research, (ii) the release of 4 different challenges that encompass control, multi-stage learning, fairness, and multi-objective learning, (iii) three baseline agents on the control setting with Proximal Policy Optimization (PPO)~\cite{schulman2017proximal}, Advantage Actor Critic (A2C)~\cite{grondman2012survey}, and Rainbow DQN~\cite{hessel2018rainbow} with a CNN state encoder, (iv) case studies of the behavior and weaknesses of the agents on all challenges. We do \textbf{not} claim that the environment allows for easy transfer of policies to real plants but we argue that this constitutes an important step towards more realistic RL environments, and supports developing agents for noisy biological settings.


\begin{figure}

\begin{center}
\includegraphics[width=0.9\linewidth]{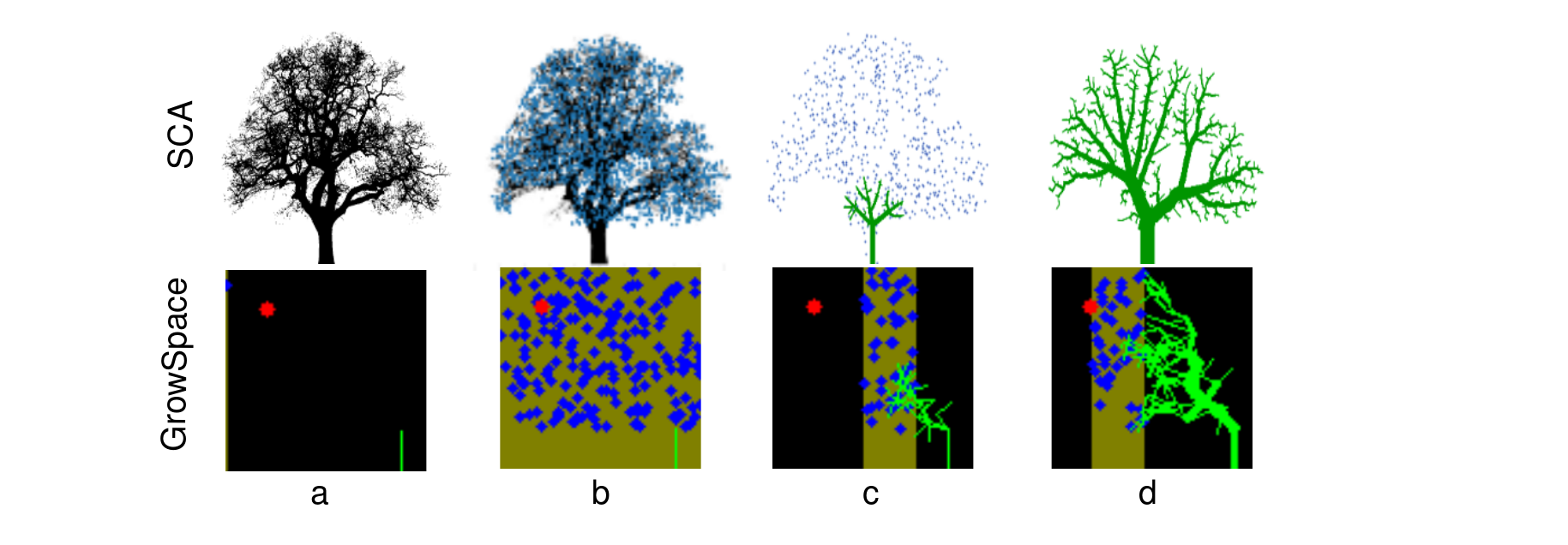}
\caption{\textbf{High-level Overview} of the approach taken for designing the \growspace{} Environment. (a) Determine target. For SCA it is the crown shape of a tree and GrowSpace it is the red circle. (b) Define a point cloud to reach target, SCA scattering surrounds tree crown and GrowSpace is stochastic. In (c) and (d) we show two different growth iterations for SCA and GrowSpace.}
\label{overviewfig}
\end{center}

\end{figure}


\section{Background}
We first cover the RL framework of a Markov Decision Process (MDP), learning with fairness constraints, and learning multiple-objectives. These topics are reviewed to lay the foundation of \growspace{} and the different challenges it provides to the RL community.
\subsection{Markov Decision Process}
A MDP is a framework used to study the control of sequential decision processes for dynamic systems~\cite{puterman2014markov}. A MDP is represented as a tuple $\mathcal M = \langle \mathcal S, \mathcal A,\mathcal R,\mathcal P, \gamma \rangle  $ that includes a state space $\mathcal S$, an action space $\mathcal A$, a transition function $\mathcal P: \mathcal S \times \mathcal A \mapsto \mathcal S$, a reward function $r: \mathcal S \times \mathcal A \mapsto \mathbb R$, and a scalar discount factor $\gamma$. For each time step $t$, a RL agent is in a state $s_t\in\mathcal S$, interacts with the environment and chooses an action $a_t\in\mathcal A$ which leads to a reward $r_t\sim r(s_t, a_t)$ and transitions to a new state $s_{t+1}\sim\mathcal P(s_t, a_t)$. The goal of a RL agent is to learn a policy $\pi: \mathcal S \times \mathcal A \mapsto [0,1]$ such as to maximize the discounted sum of rewards.

\subsection{Fairness in RL}\label{fairness_secion}

Fairness is of concern in RL when actions selected by the agent affect the state and latter rewards. In a MDP setting, several constraints of fairness have been introduced over the past years. In the multi-armed bandit learning framework, fairness has been studied in the setting where the selection of an arm with lower expected reward over another arm is considered unfair~\cite{joseph2016fairness}. \textcite{jabbari2017fairness} implement this constraint in an MDP setting, stipulating that in any state $s$, an algorithm cannot favor action $a$ that has a lower probability of a expected reward than action $a'$. For the scope of this paper we select a fairness constraint widely used in a supervised learning setting \cite{hardt2016equality} and which has been evaluated in a MDP setting in reinforcement learning by  \textcite{wen2019fairness}. The fairness constraint is to provide equality of opportunity and achieve demographic parity amongst $N$ growing plants. This is quantified through the reward to the agent (in our case light moving agent) where the rewards reflects the quality of treatment of $N$ plants.

\subsection{Multi-objective RL}

Multi-objective reinforcement learning involves having two or more objectives that may conflict with each other and need to be achieved by an agent~\cite{vamplew2011empirical}.  Rewards in this context are a feedback signal that are represented as a vector of length equivalent to the number of objectives to attain~\cite{brys2014multi}. Conflicts amongst objectives are observed when certain objectives are being favored over others. To reduce conflicts, trade-offs are used between objectives. The most widely used optimally criterion is the Pareto dominance relation~\cite{van2014multi}. Pareto dominance happens at the policy level, when a policy surpasses all other policies for all objectives. Learning policies that meet all preferences has been shown to be a challenging task and consequently the problem is often reformulated as a single-objective problem in the literature~\cite{yang2019generalized}. This comes with limitations because certain behaviours can emerge and show preferences to one of the objectives.


\section{Related Work}

The proposed \growspace{} environment complements current RL benchmarks and existing plant modelling platforms.

\subsection{RL Benchmarks}

The Arcade Learning Environment (ALE)~\cite{bellemare2013arcade} has long been used as a benchmark for evaluating AI agents on a variety of tasks. These tasks have pushed our knowledge and the direction of research notably in representation learning, exploration, transfer learning, model learning, and off-policy learning~\cite{machado2018revisiting}. Similarly, StarCraft II~\cite{vinyals2017starcraft} presents harder tasks than prior video game-based environment. However, as mentioned earlier the usage of common benchmarks has been put into question and how they could translate to the real world~\cite{henderson2018deep}. Recently, interest has been pushed on procedurally generated environments such as Procgen Benchmark~\cite{cobbe2020leveraging} and the NetHack Learning Environment~\cite{kuttler2020nethack} both with the intent of tackling generalization with large amount of tasks and levels. The focus of these benchmarks are not real world-orientated.
The closest RL benchmark to real-world interaction is Mujoco~\cite{todorov2012mujoco}, a physics engine that enables testing of robotic simulations with contacts. Although Mujoco can adapt different types of bodies and movements, no task formulation has been addressing a greater challenge such as fairness. \growspace{} fills this gap.

\subsection{Plant Modeling}

Plants are interesting subjects to simulate as they are self-organizing systems that have the ability to adapt to dynamic environments by sensing their surroundings and directing their growth to preferable regions~\cite{calvo2011plants}. Plant models have evolved throughout the past two decades and have been incorporating the effects of environmental conditions~\cite{boudon2012py}. Simulation of realistic virtual plants and trees have been explored through different algorithms such as L-systems~\cite{prusinkiewicz2002art}, Functional–Structural Models (FSMs)~\cite{de2003relevant} and Space Colonization Algorithms (SCA)~\cite{runions2007modeling}. Plant modeling has received increased interest and has primarily focused on: the reconstruction of plant architectures overtime, discovery of underlying ecophysiological mechanisms driving certain plant traits, and the movement of nutrients and their allocation throughout the plant body~\cite{evers2018computational}. The development of these models are beneficial and accelerate the ability to understand the functioning, manipulation and provide a way to test out hypotheses on plant growth. However, they are not feasible for generating and controlling behavioral patterns that a plant may exhibit~\cite{bonnet2019robots}. We are basing our simulator on the SCA, adding a controllable light source and target points and shapes for the plants to grow towards.


\section{\growspace{} Learning Environment}

We present \growspace{}, a new procedurally generated RL environment that is built with the OpenAI gym interface~\cite{brockman2016openai}. The simulator is inspired by a real-world problem of optimizing plant physiology and direction of growth over time. In the real-world, plant growth is dictated by several variables, an important one is light availability. \growspace{} incorporates a plant's behavioral response to light and provides control over the branching by means of a mobile light (either light beam or small spotlight). The objective is to guide the growing plant to a desired target or shape depending on the challenge.
Figure~\ref{overviewfig} provides an overview of our approach for designing \growspace{}. Much like in the real world, the light directly influences the direction of growth of a plant. A branching algorithm is chosen to mimic a plant's relationship to light. Finally, the branching algorithm is formulated as a RL problem where an agent's objective is to shape a plant towards a target (red) or a desired configuration through means of a mobile light.

\subsection{Plant Branching}\label{plant branching}

The SCA ~\cite{runions2007modeling} model is chosen for \growspace{} due to its success in generating virtual trees with features that are comparable to real trees\cite{yandun2020visual}. Through the attachment of plant segments to a plant structure, this algorithm facilitates the iterative growth of a virtual plant. 
\begin{SCfigure}
  \caption{Steps for branching in the Space Colonization Algorithm, where (a) all photons are filtered (b) trough a radius of attraction (c) and their normalized vectors from the plant tip to the photons (d) are summed and normalized to find the direction of growth (e) for the new plant segment to be attached (f) process is repeated for all existing plant tips}
  \includegraphics[width=0.5\textwidth]{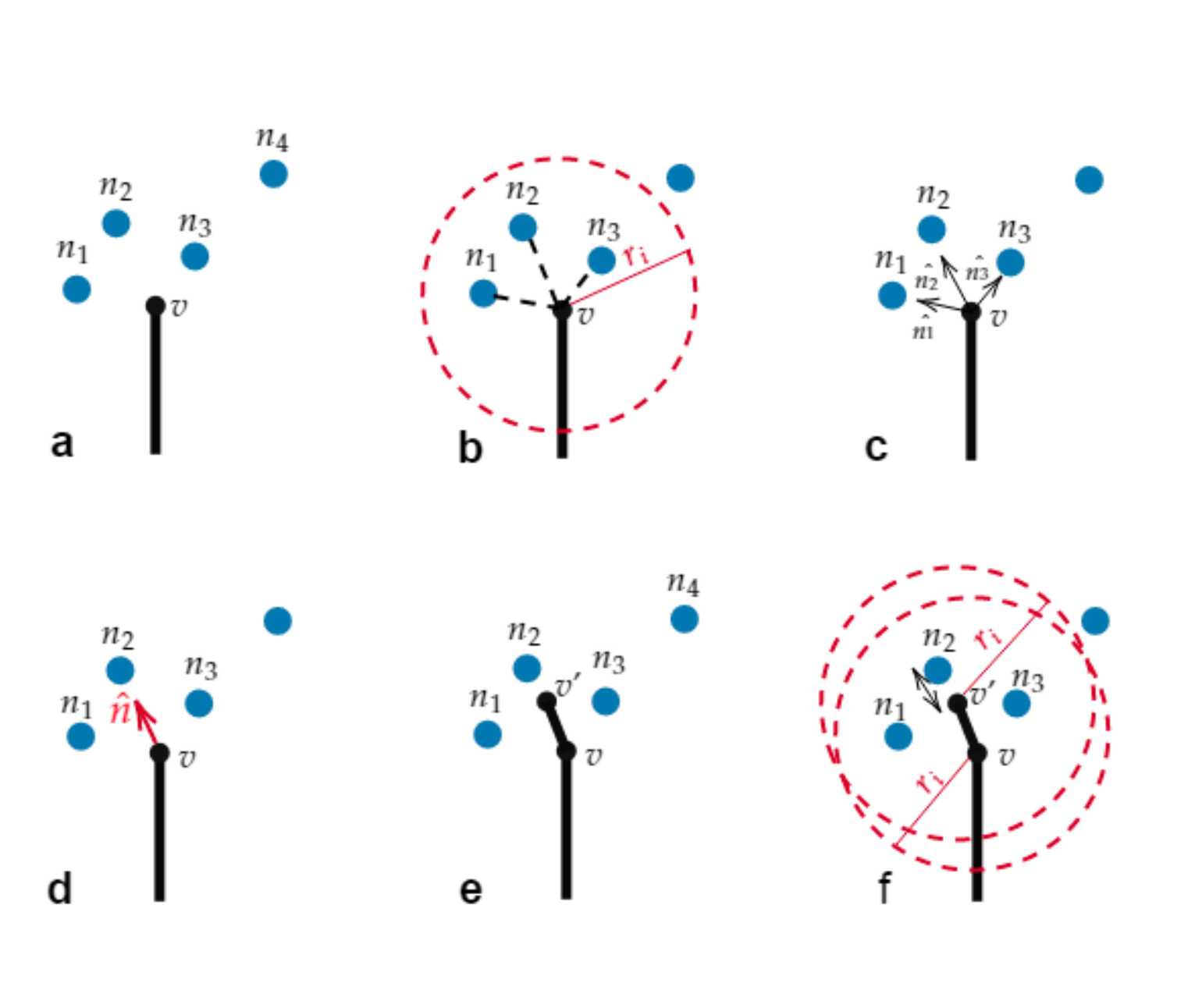}
  \label{figure2electricfigaloo}
\end{SCfigure}

In Figure~\ref{figure2electricfigaloo} (inspired by~\cite{runions2007modeling}) illustrates the algorithm. The algorithm begins with a set of photon particles $N$ and an initial plant segment with tip $v$ (a). The plant segment tip eventually become a set as the plant grows, where $v \in {V}$. In order for a plant tip to grow, photons $n \in N$ must be located within a predefined radius of influence $r_i$, as seen in (b) where $n1$, $n2$ and $n3$ attract segment tip $v$. When a photon is too close to a plant segment, the photon is removed and is not considered. The normalized vectors from tip $v$ towards photons $n \in N$ are computed (c). Once summed, the normalized vector $\hat{n}$ is found for $v$ (d). The vectors representing the direction of growth are:
\begin{align}
    \overrightarrow{n} = \sum_{n}^{N} \frac{n-v}{\|n-v\|}.
\end{align}
The final normalized vector for a plant segment tip is:
\begin{align}
    \hat{n} = \frac{\overrightarrow{n}}{\| \overrightarrow{n}\|}.
\end{align}

Vector $\hat{n}$ is the direction of growth of plant segment tip $v$ and is towards photon $n_2$. The plant grows a new segment $v'$ (e). The procedure is then repeated on both of the plant segment tips. In (f) we can observe that $n_2$ is too close to $v'$ and will not be considered for branching.

In \growspace{} the attraction points are thought of as available photon particles and are scattered at random to facilitate stochastic branching. The amount of observable photons are limited and available to the plant with a light source. The light source illuminates photons within a certain range (that are not visible to the agent) and consequently restricts the direction of growth. In Figure~\ref{overviewfig} we enable the scattering of photons to be visible for better understanding of \growspace{}. We introduce the concept of light direction to encourage unidirectional growth towards the light source. To grow towards the light source, shading needs to take place as to not allow the light beam to illuminate the photons that are below existing parts of the plant foliage. This integration is based on phototropism, a response process, that enables plants to adjust their growth towards the direction of the light~\cite{goyal2013phototropism}. Further details about SCA can be found in the Appendix ~\ref{add_grow}. 

\subsection{Reinforcement Learning Framework}

We formulate \growspace{} as a MDP described by a state space $\mathcal S$ that is accessed by the agent as a pixel observation, an action space $\mathcal A$ that can be discrete or continuous, a transition function $\mathcal P$ and a reward function $R$. On each time step $t$ of a learning episode, the agent observes the state $s_t\in\mathcal S$, takes an action $a_t\in\mathcal A$, moves to a new state $s_{t+1}\sim \mathcal P(s_t, a_t)$, and receives a reward $r_{t+1} \sim R(s_t, a_t, s_{t+1})$. The probability of a plant segments tips to branch in a specific direction given action $a_t$ in state $s_t$ is incorporated into the transition probability $P(s_{t+1} \rvert s_t, a_t)$. In this environment, much like in the real world, the light directly influences the direction of growth of a plant. The agent's objective is to shape a plant towards a target or a desired configuration through means of a mobile and adjustable light source.

\textbf{States and Observations:}
For every step taken in the environment, the agent observes the \emph{observation} of its current \emph{state} prior to selecting an action. Once an action is selected by the agent, the new state becomes the observation for the next time step. States and observations are an image representation of the environment which display the plant structure, the light source and the target.  The observations are available to the agent as an RGB image that contains the plant, the target and the light beam at time step $t$. The dimensions are of of 84  $\times$ 84 $\times$ 3 , except for the plant shaping challenge where the dimensions are of 28 $\times$ 28 $\times$ 3. 

\textbf{Actions:}\label{actionspace}
\growspace{} provides a discrete action space and a continuous action space. In the discrete action space the agent can execute five discrete actions. The agent can move the light beam to the right, the left or stay put. The agent can equally increase or decrease the available light beam to the plant. The movement of the light beam is set at a default of 5 pixels in any given direction and can be customized by the user. The continuous action space has two actions, the light velocity, speed at which the light is displaced, and the width of the light beam. This could be a more realistic and more complex set-up, and it will help to transfer the problem from simulation to real world. The actions chosen will influence the available scattering to the plant and will impact the direction of growth of the plant. For example, if the beam of light is not close enough the plant will not be able to branch out because the attraction points and will be dormant. If the light reveals several points, branching will be occur in multiple places in the illuminated area. 

In the multiple-objective task, the action set changes due to the circular light beam. Similarly, the agent can increase or decrease the light beam radius, it can move left and right and, can move up and down. The default radius of the beam is 10\% of the width of the environment.

\textbf{Rewards:}
The reward will be dense and will be received at each time step. Rewards will depend on the challenges in which the agent is trying to solve. Rewards are task specific and explained below in Section~\ref{tasks}.

\textbf{Episode and Reset:}
The episode length is fixed and is set to 50 steps. At the beginning of each episode, the scattering of photons, and the initial plant stem, as in Section~\ref{plant branching}, and the target(s) are procedurally generated in order to ensure the agent will not have visited the exact state previously in other episodes.


\section{Tasks}\label{tasks}

We propose an initial set of tasks that can be tackled in \growspace{}, all of which with several levels of difficulty. The combination of tasks released encompass some known challenges to the RL community, such as control, multi-stage learning, fairness, and multi-objective learning. 

 
 


\textbf{Grow Plant to Goal:} The task consists in growing the plant with the light beam towards a target positioned at random in the upper 25\% of the environment. Every episode begins with the light beam positioned above the original plant stem. The agent must displace the light beam to control and direct the growth of the plant towards the target. After each action, the agent is rewarded based on the smallest distance between any of the branch tips and the target. Let $d_{b,g}$ denote the Euclidean distance between a branch tip $b$ and a target goal $g$:
\begin{align}
   d_{b,g} = \sqrt{(x_b - x_g )^2 + (y_b - y_g )^2}.
\end{align}
The reward obtained at time step $t$ is inversely proportional to this distance of the  branch tip closest to the goal among the current branch tips $\mathcal B_t$:
\begin{align}
    R_t = \max_{b \in\mathcal B_t} \frac{1}{d_{b,g}} \label{eq:control_reward}.
\end{align}
Rewards are therefore in the range $]0, 1[$. This typical control problem~\cite{recht2019tour} is considered the simplest of the tasks since the light movements directly impact the plant from the beginning of the episode. The difficulty of this task is proportional to the distance between the target and the original plant stem tip; as the distance increases, the harder the task becomes. 

\textbf{Find Plant:} The task consists in finding the original plant stem with the light source, either the beam or circular light. An episode starts with the light source and the original plant stem positioned at different random locations in the environment. This becomes a multi-stage learning problem~\cite{ni2017reinforcement} where the agent has to first locate the original plant stem by displacing the light source in order to increase the reward signal. The reward is computed using Equation~\ref{eq:control_reward}.
The difficulty of this task is proportional to the distance between the target and the original plant stem tip (as in the Grow Plant task), and to the distance between the original plant stem and the initial light source position. Displacing the light source multiple times before finding the plant reduces an agent's ability to attain the highest amount of rewards.

\textbf{Grow Multiple Plants:} The task consists in finding two or more plant stems with the light beam and growing them to similar maturities throughout the episode. In this task, the agent must grow $n > 1$ plants towards a target. The target is placed at random in the upper 25\% of the environment, the light beam and initial plant stems are initialized randomly within the environment. As in the Find Plant task, the agent must displace the light beam to find all the existing plants in order to initiate a reward signal. Based off our chosen fairness constraint mentioned in section \ref{fairness_secion} where the objective is to provide equal opportunity and achieve parity amongst plants, the reward signal is crafted as the minimum distance reward (Eq.~\ref{eq:control_reward}) over all plants:  
\begin{align}
    R_t = \min_{1 \leq i \leq n} R_t^{(i)}, \label{eq:3}
\end{align}
where $R_t^{(i)}$ is the grow plant reward (Eq.~\ref{eq:control_reward}) associated with plant $1 \leq i \leq n$.
As seen in Section~\ref{fairness_secion}, different fairness constraints can be adopted in a MDP setting and could be integrated within \growspace{}. We set our first fairness task with a fairness constraint that is similar to~\cite{wen2019fairness}, which suggests that the agent should provide equal opportunity for each plant to grow towards the target at every step of the episode. The difficulty of this task is in sharing the amount of available photons adequately between plants when they start growing closely to each other. As different plants start approaching each other the photons may run out in the desired direction of the target and the plants may never reach the target (see Appendix ~\ref{videoagent}).

\textbf{Grow Plant to Shape:} This task consists in growing plants into specific shapes by using a circular light source that can navigate to precise locations in the environment. As default shapes for benchmarking purposes we consider the MNIST dataset~\cite{lecun1998mnist}, which is widely used in machine learning. MNIST contains $28 \times 28$ pixel binary images of  handwritten digits (0-9). Given an MNIST image, the goal is to grow a plant such that its shape matches the drawn digit as best as possible. For this task, the environment is reshaped to a width and height of $28 \times 28$ pixels (i.e. the size of a MNIST image). The agent has to grow the plant into multiple directions to best cover the outline of the MNIST digit without growing out of bounds. This is a multi-objective task, since the agent has to cover multiple areas in any order, while also keeping the overall goal of limiting the amount of branching in mind.

The reward for this task is crafted using the Jaccard Index~\cite{fletcher2018comparing} similarity score. Let $\mathcal A_t$ and $\mathcal G$ respectively denote the set of pixels that the plant occupies at time steps $t$ and the set of pixels that belong to the target shape. The reward at time step $t$ is given by the similarity score:
\begin{align}
   R_t = \frac{\mathcal A_t\cap \mathcal G}{\mathcal A_t \cup \mathcal G}. 
\end{align}


\section{Experiments and Results}\label{experimentsresults}

In this section we demonstrate how \growspace{} is challenging by comparing RL baseline algorithms and through case studies on different settings of the available challenges.

\subsection{Baselines:}

We present the different gradient-based policy methods we have evaluated on the easiest challenge in \growspace{}, the control challenge (Grow Plant to Goal task). We compare Proximal Policy Optimization (PPO)~\cite{schulman2017proximal}, Advantage Actor Critic (A2C)~\cite{grondman2012survey}, and Rainbow DQN~\cite{hessel2018rainbow}. For each of these agents, a state is represented by a tensor of $(3, w, h)$ where $w$ and $h$ are the width and height of the observed image in the task. These representations are fed through three convolutional layers, a fully connected layer and a final layer using the ReLU activation function. The output of the policy network is a probability of each action belonging to the action space. The mean episodic reward over 1 million steps is used to display the learning behaviour of the different RL agents. 

\begin{figure}[h!]
\begin{center}
\includegraphics[width=0.4\columnwidth]{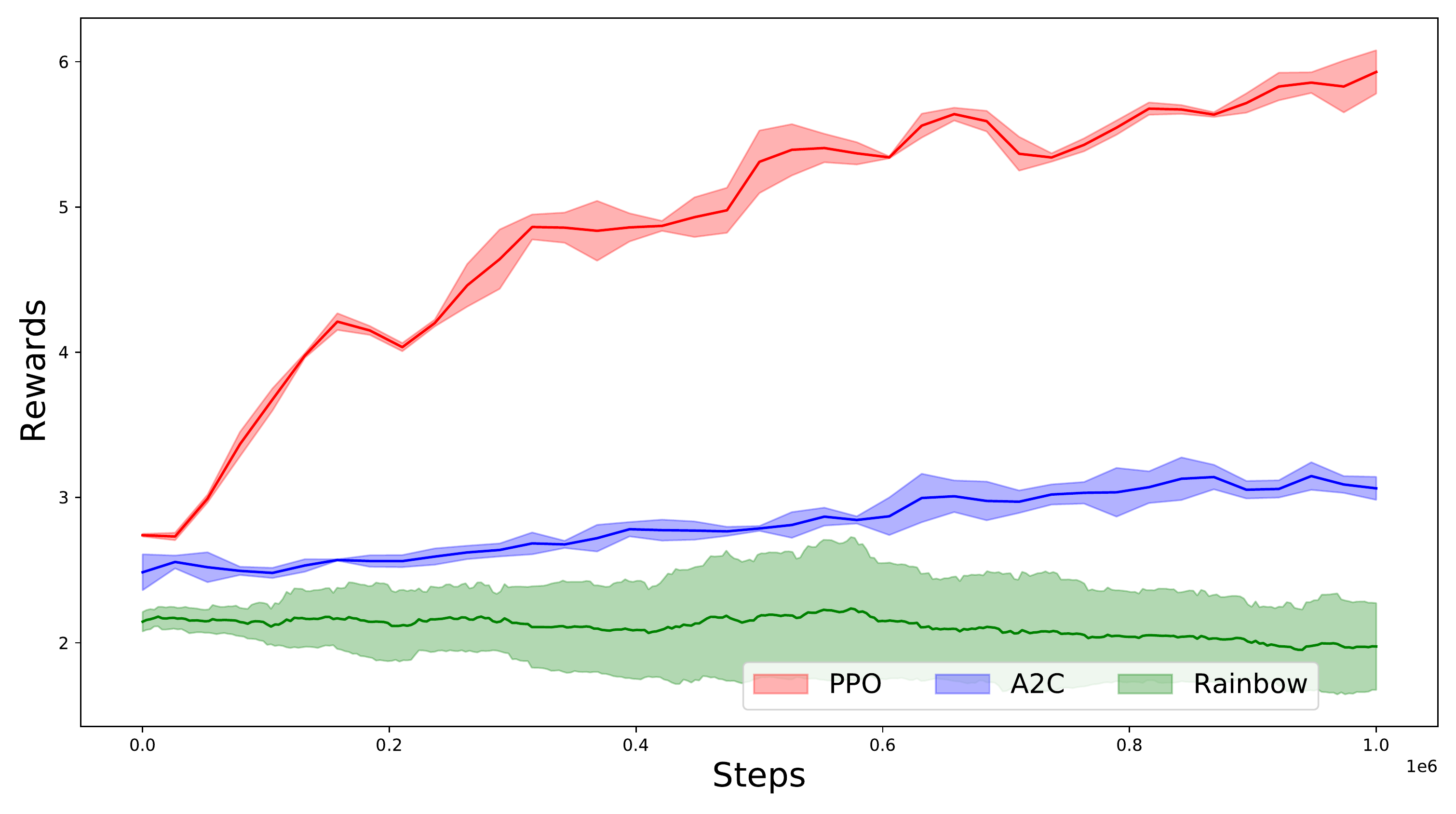}
\caption{Baseline comparisons for the control challenge}
\label{fig:control-baselines}
\end{center}
\end{figure}

In Figure ~\ref{fig:control-baselines} we can observe that PPO displays a greater learning behaviour compared to A2C and Rainbow DQN. Based off these results, PPO was the most promising strategy and was selected as our main baseline to conduct our case studies. We therefore conducted a hyperparameter search for PPO across all challenges and with three different seeds. The details of the final chosen PPO parameters can be found in Appendix~\ref{appendix_b}. We equally provide computing times in Appendix~\ref{compute} to demonstrate that \growspace{} is a fast running benchmark, similar to Atari. 

\subsection{Case Studies}\label{casestudies}

\begin{figure}[h!]
\begin{center}
\includegraphics[width=\columnwidth]{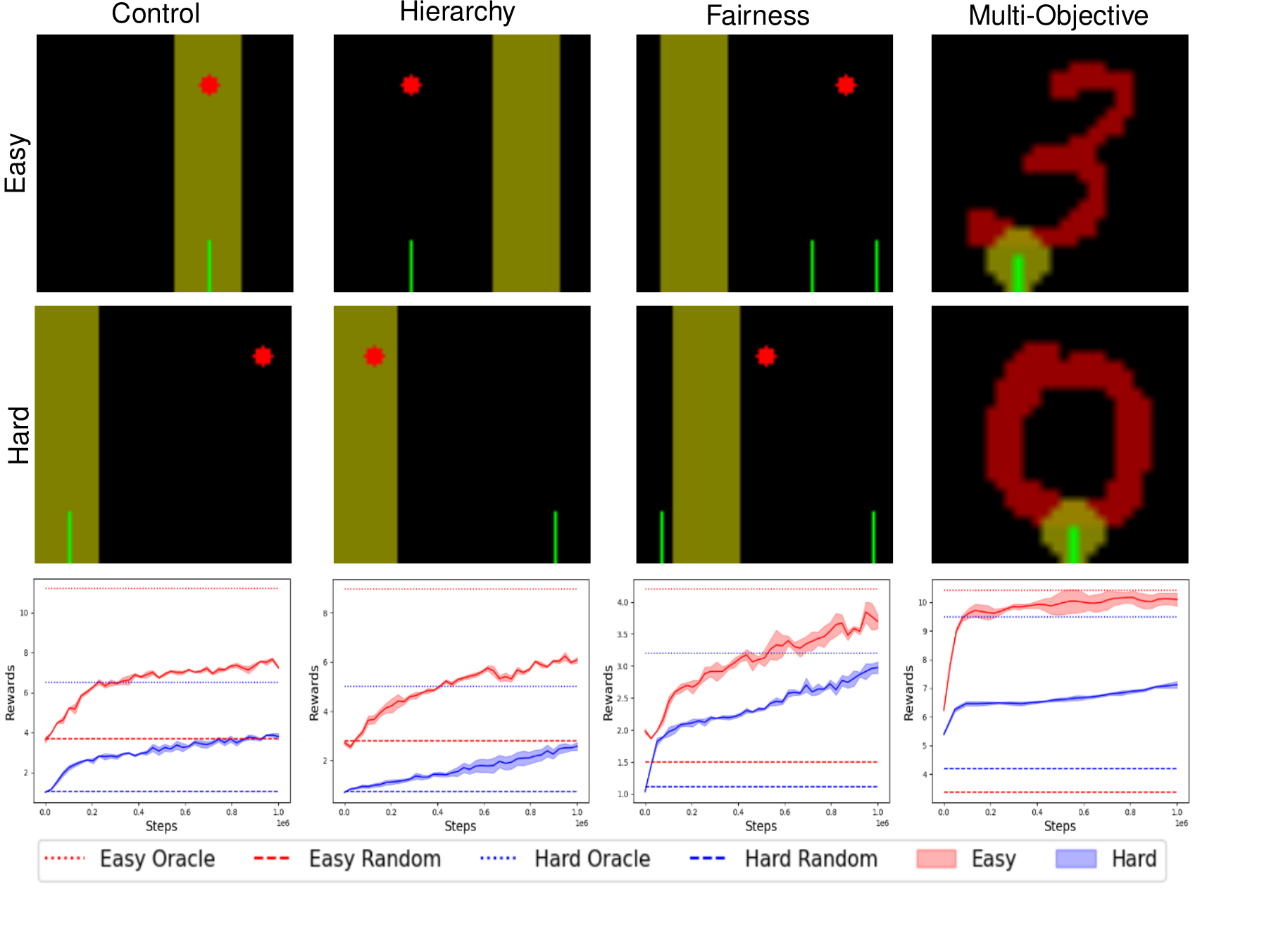}
\caption{Episode initialization for all case studies}
\label{fig:all_case_studies}
\end{center}

\end{figure}

We present a set of case studies with our main baseline PPO \cite{schulman2017proximal} to showcase the spectrum of behaviors of the agent and where challenges are shown to be difficult. For each challenge, we provide an easy and a hard case study, to be described below. Figure~\ref{fig:all_case_studies} displays the initialization of an episode for all challenges in easy and hard settings as well as the the cumulative rewards (averaged and one standard deviation). To better understand the performance of PPO, a random agent and an oracle agent have been implemented for each challenge. No training was performed for the random and oracle agents. The random agent selects actions uniformly at random from the action space. More information about the oracle solutions can be found in Appendix~\ref{add_case}. The mean episodic reward is our performance metric to evaluate if learning is successful. We include other metrics such as the selection of actions and the overall number of branches produced throughout an episode, these can be found in Appendix ~\ref{add_grow}.Results are always averaged over three runs (different random seeds). 

\textbf{Control:}
The easy setting is when the target is above the original plant stem and the hard setting is when the stem and target are at opposite extremities of the environment. Figure~\ref{fig:all_case_studies} shows that the easy control reward curve from PPO is closer to the oracle solution and that learning can be improved. For both settings we observe that the PPO reward curve is midpoint between the oracle and random action selection, suggesting that PPO's behaviour can be optimized further. The video renderings show the agent displacing the light away from the plant too quickly, loosing steps with stagnating rewards instead of growing new closer branches and results in guiding the plant to target. The episodic action selection as seen in Figure ~\ref{Control easy} in Appendix ~\ref{add_case} demonstrates that the agent does not favor decreasing the light beam resulting in a plant with multiple branches competing for the same photons in the direction of the target and thus resulting in slower growth and lower rewards. The action distribution in the easy setting is relatively similar amongst actions, however in the hard setting it is noticeable that the right and left actions are used more (see Figure ~\ref{fig:fig_control_case} in Appendix ~\ref{add_case}). This can be explained as the plant needs to grow vertically and laterally to the opposite side of the environment.

\textbf{Multi-stage Learning:}
Similar to control case study settings with the exception that the initialization of the episode starts with the light placed at random. Figure ~\ref{fig:all_case_studies} shows that the hard setting has a lower reward due to the distance between the initial plant stem and the target. With the initial task of finding the plant first, the low reward in the hard setting can be explained by the agent receiving the same reward while trying to find the plant and, the greater distance between the target and the initial stem. The action of increasing the light is more utilized within the harder setting to find the initial plant stem (see Figure~\ref{fig:fig_hierarchy_case} in Appendix~\ref{add_case}). With the video renderings, we the light width does not change a lot once the initial stem is found and the agent learns to drag the light towards the target. The video renderings show that the plant gets bushy and the smaller light width is not utilized efficiently to try and reduce competition amongst branches for available photons (see Appendix ~\ref{videoagent}).

\textbf{Fairness:}
The easy setting starts with the episode initialization with both plants at a distance that is set to the default light width and the target is placed in the middle. For the hard setting, the initialization of an episode starts with the plants at the opposite extremities of the environment and the target is placed in the middle of the environment.  This case study is particular because the plants are very close and competing for available photons in order to reach the target. As a fairness challenge, the objective is to produce plants of similar size. Figure~\ref{fig:all_case_studies} shows that the easy fairness reward curve from PPO produces the highest amount of rewards. Both PPO reward curves are between closer to the oracle bound than the random agents for both cases. We investigate if the agent's behaviour is fair by looking at the median amount of branches per plant, where the numbers are relatively close (see Figure~\ref{fig:fairbranches} in Appendix ~\ref{add_case}. The easy case produces a smaller amount of branches, this can be explained by the small pool of photons that are available to both plants branching and thus limiting additional branching in the right direction. In the middle case, the branching is higher and can be explained by the greater amount of photons available to both plants while reaching the target as they do not need to compete for the majority of the episode.

\begin{figure}[h!]
\centering
\subfigure[MNIST digit comparison]{%
    \includegraphics[width=0.4\columnwidth]{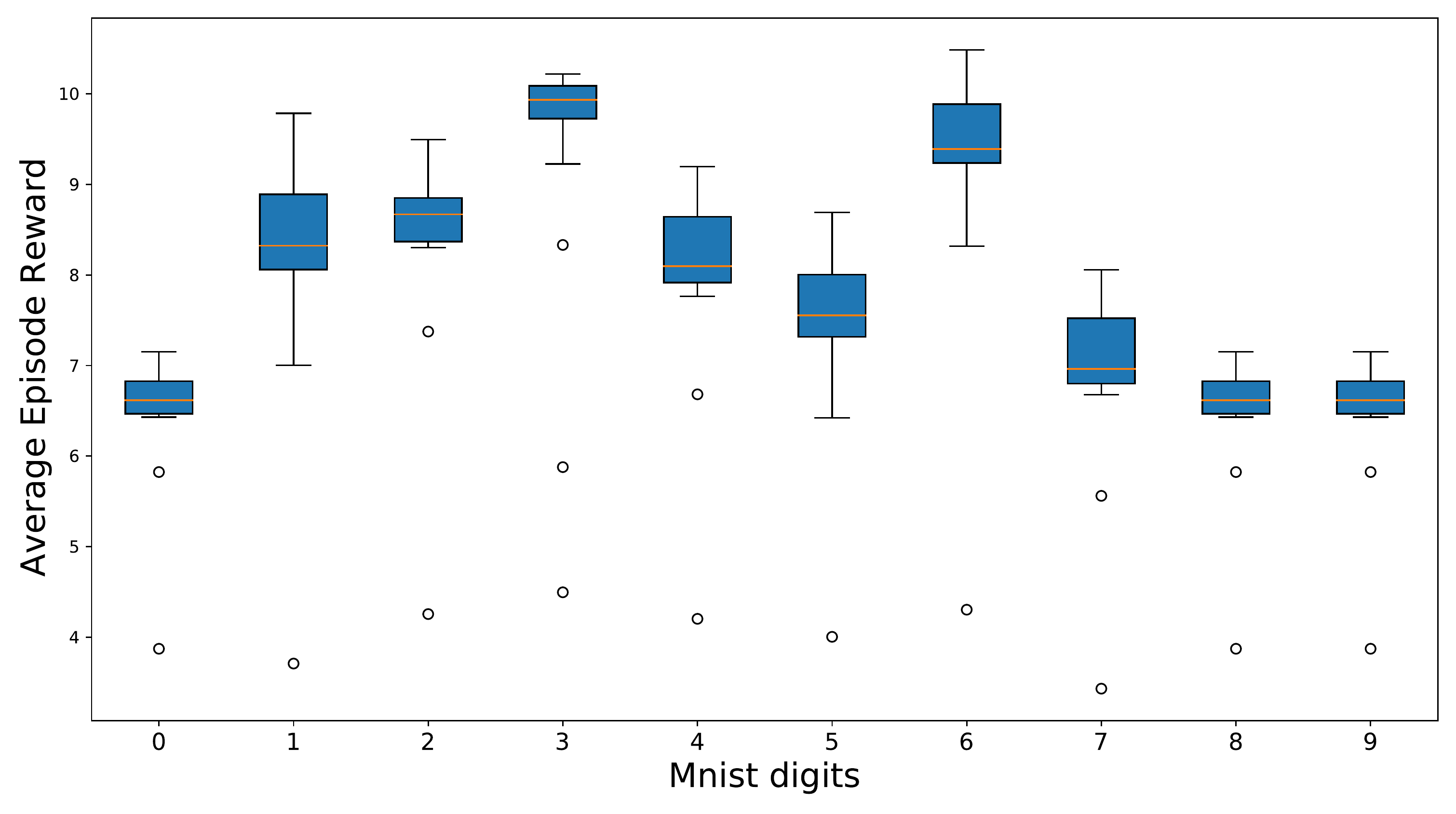}
    \label{mnistdigit}}
\quad
\subfigure[Curriculum comparison]{%
    \includegraphics[width=0.4\columnwidth]{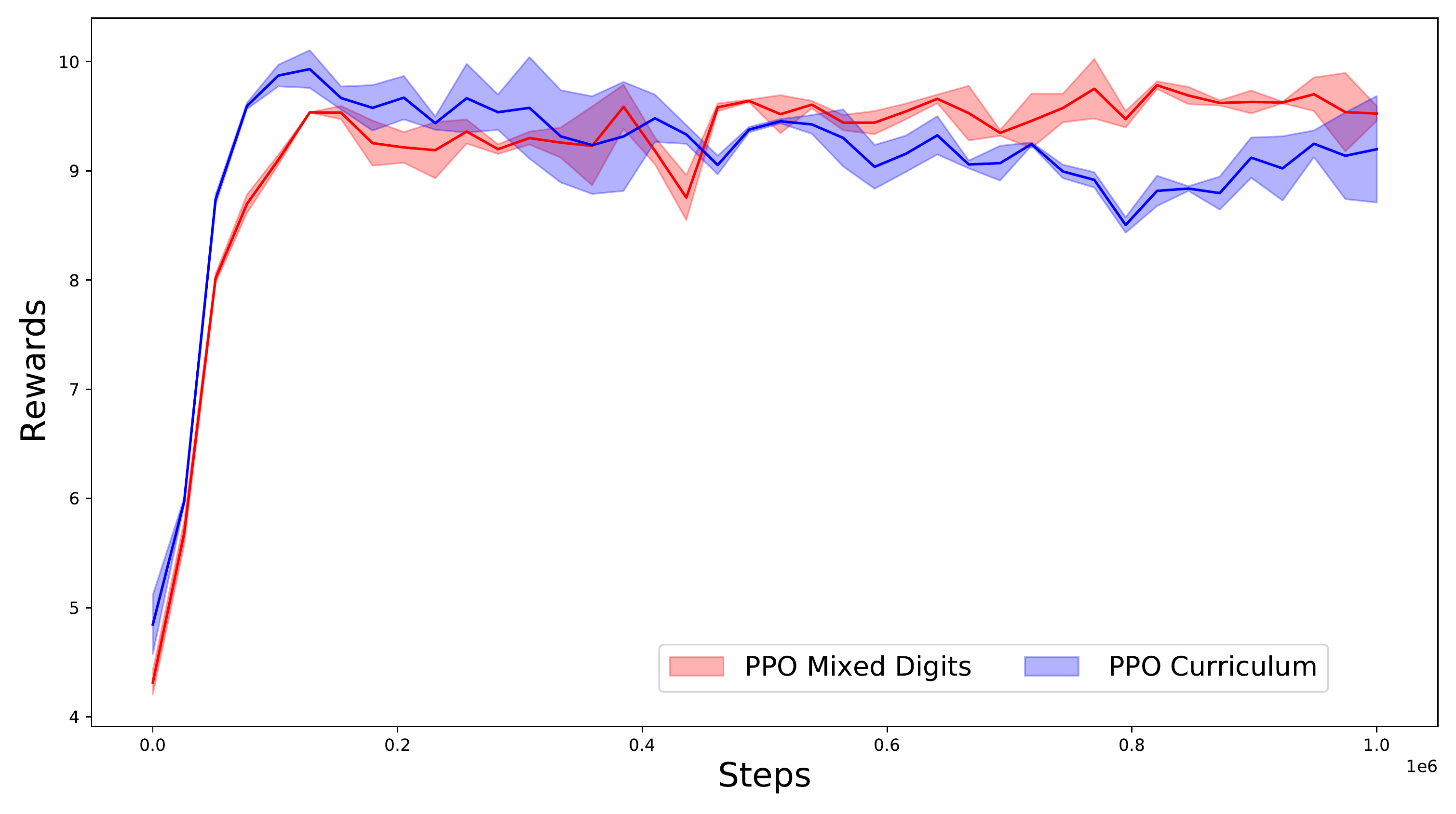}
    \label{curriculum}}
\caption{Comparison of digits to design the curriculum for training}
\label{fig:mnistcases}
\end{figure}

\textbf{Multi-objective Learning:}
The easy settings consists of growing a plant in the shape of digit 3. For the hard setting, the agent needs to grow in the shape of digit 0. The episode initialization for respective case studies starts with the tip of the initial stem touching the MNIST digit and the spotlight placed at random. The easy and hard digits were determined by running PPO individually on all MNIST digits separately. The digits are compared by their median reward values from PPO as seen in Figure\ref{mnistdigit} and are ranked from easiest to hardest respectively: 3, 6, 2, 1, 4, 5, 7, 8, 9, 0 . Figure~\ref{fig:all_case_studies} shows that the easy multi-objective reward curve from PPO produces the highest amount of rewards. The hard multi-objective challenge is indeed more difficult as the highest reward achieved by the oracle is lower than in the easy setting and is midpoint between the oracle and the random curves. The video renderings show the agent never fully succeeding at filling in the easy digit and the hard digit suggesting that PPO's behaviour and the can be optimized further (see Appendix ~\ref{videoagent}). This highlights that the multi-objective challenge is difficult and that there is room for improvement for the oracle strategy.  

An additional case study was explored with the ranked MNIST digits. A curriculum was constructed starting with the easiest digits and consists of 2000 episodes with the two first easiest digits and for every increment of 2000 episodes a new new digit is added. The last 6000 episodes of training have all the MNIST digits.
In Figure \ref{curriculum} the learning seems at a higher rate in the first episodes of training for the curriculum approach however, the reward curve decreases as the addition digits are added. The random selection of digits seems to be a better fit over time. We can see that the agent is focused on density on plant vs overall shape as the light width fluctuates a lot in the video renderings but it does not visit the full trajectories of the MNIST digits.


\section{Limitations}
The limitations of \growspace{} are translating plant growth control into practice. The benchmark provides a modest  first step to modeling a plant response that occurs in the physical world however, under the assumption of all other environmental conditions being constant (water supply, wind, nutrient availability, etc). The transfer of an optimal policy in simulation may not succeed when reproducing the experiment in the real-world however, high-level intuition can be extracted from the optimal policies \cite{christiano2016transfer}. \growspace{} implements one plant model for a generalized perspective into plant growth, specific models for different plant species could enable better precision and simulations that are specific to researchers needs.


\section{Conclusion and Future Work}
\growspace{} is a procedurally generated environment with a set of challenges that can help the advancement of reasearch in RL and agriculture. It encompasses real-world behaviour of plants in a low representation setting and provides a series of challenges that address issues such as fairness. We provide gradient based agent baselines for the control challenge to display the difficulty of the easiest challenge within \growspace{}. Case studies with our base performing baseline, PPO, are layed out to give insights on the type of behaviour an agent can adopt in easy and hard settings. We demonstrate that indeed \growspace{} is a environment that is complex and proposes different settings which enable different skills to be learnt such as sharing ressources in the fairness constraint, patience for displacing a light to grow the plant and limiting available resources to a growing plant for precision. We illustrate that the challenges within \growspace{} are hard with PPO in some case studies only performing at 50\% human level.

Further add-ons can be attainable in order to recreate a full growing environment dynamic with water, nutrients, wind and even specific plant models. We plant to support \growspace{} after its release as well as introduce new environment parameters. In sum, plant growth is a grounded and intricate topic and its full control is not fully understood. \growspace{} provides a first step in the direction of plant growth control through a known plant response, phototropism.


\newpage

\printbibliography



\newpage
\section*{Checklist}


\begin{enumerate}

\item For all authors...
\begin{enumerate}
  \item Do the main claims made in the abstract and introduction accurately reflect the paper's contributions and scope?
    \answerYes{}
  \item Did you describe the limitations of your work?
    \answerYes{}
  \item Did you discuss any potential negative societal impacts of your work?
    \answerNo{}
    Although this is based on a real plant growth simulator, if they derive a system based on this they may not get what they are thinking
  \item Have you read the ethics review guidelines and ensured that your paper conforms to them?
    \answerYes{}
\end{enumerate}

\item If you are including theoretical results...
We do not include theoretical results in this paper.
\begin{enumerate}
  \item Did you state the full set of assumptions of all theoretical results?
    \answerNo{}
	\item Did you include complete proofs of all theoretical results?
    \answerNo{}
\end{enumerate}

\item If you ran experiments...
\begin{enumerate}
  \item Did you include the code, data, and instructions needed to reproduce the main experimental results (either in the supplemental material or as a URL)?
    \answerYes{}
  \item Did you specify all the training details (e.g., data splits, hyperparameters, how they were chosen)?
    \answerYes{}
    Please refer to Appendix ~\ref{appendix_b}
	\item Did you report error bars (e.g., with respect to the random seed after running experiments multiple times)?
    \answerYes{}
	\item Did you include the total amount of compute and the type of resources used (e.g., type of GPUs, internal cluster, or cloud provider)?
    \answerYes{}
    Please refer to Appendix ~\ref{compute}
\end{enumerate}

\item If you are using existing assets (e.g., code, data, models) or curating/releasing new assets...
\begin{enumerate}
  \item If your work uses existing assets, did you cite the creators?
    \answerYes{}
 We have cited the branching algorthim, Section ~\ref{plant branching}, and all baselines used for building and testing \growspace{} Section ~\ref{experimentsresults}.
  \item Did you mention the license of the assets?
    \answerYes{}
    We include the MNIST dataset that is properly cited in Section ~\ref{tasks}.
  \item Did you include any new assets either in the supplemental material or as a URL?
    \answerYes{}
    Access to \growspace{} is provided via URL and is shared in the supplemental information.
  \item Did you discuss whether and how consent was obtained from people whose data you're using/curating?
    \answerNo{}
In this paper we do not utilize data from other people. We cite accordingly for algorithms used.
  \item Did you discuss whether the data you are using/curating contains personally identifiable information or offensive content?
    \answerNo{}
    Not applicable to \growspace{}.
\end{enumerate}

\item If you used crowdsourcing or conducted research with human subjects...
\begin{enumerate}
  \item Did you include the full text of instructions given to participants and screenshots, if applicable?
    \answerNo{}
    Not applicable to \growspace{}.
  \item Did you describe any potential participant risks, with links to Institutional Review Board (IRB) approvals, if applicable?
    \answerNo{}
    Not applicable to \growspace{}.
  \item Did you include the estimated hourly wage paid to participants and the total amount spent on participant compensation?
    \answerNo{}
    Not applicable to \growspace{}.
\end{enumerate}

\end{enumerate}


\newpage
\appendix
\section{Additional Details on \growspace{}}\label{add_grow}

\subsection{Parameters} \label{grow_param}

In Table ~\ref{tab:growparam} we describe how  parameters of \growspace{} can influence the overall output shape of a growing plant. Default values are provided but can be changed by the user.
\begin{center}
\begin{table}[h]
\centering
\begin{tabular}{p{0.2\columnwidth} p{0.8\columnwidth}}
\hline
 \multicolumn{2}{c}{\textbf{\growspace{} Parameters}} \\
 \hline
 \small \vfil \hfil Light Density & Increases stochastic branching and chance of branching as in increases. If too small a plant may not grow due to a photon never being in the radius of attraction\\
 \hline
 \small \vfil \hfil First Branch Height & This will reduce the amount of branching if increased and facilitates reaching photons that are closer to the target. If decreased, the plant can become bushy quickly and may never attain target when light width is not controlled properly.\\
 \hline
 \small \vfil \hfil Max Branching & Upper limit of possible amount of new branches per step, the greater it is the bushier the plant will get, the lower it is, the chance of branching reduces\\
 \hline
 \small \vfil \hfil Branch Length & Length at which every branch can grow. If value is high this results in faster growth. If value is low the plant will grow more slowly and will need more steps to attain the same target. \\
  \hline
 \small \vfil \hfil Light Displacement & Increment at which the beam or focal light can move in any direction. If too big it may skip photons along a desire trajectory, if too small, this may lead to not completing task. \\
\hline
\end{tabular}
\caption{Parameters that can be changed by the users in \growspace{}}
\label{tab:growparam}
\end{table} 
\end{center}


\section{Hyperparameter Search}\label{appendix_b}
A first hyperparameter search was performed across \growspace{} parameters that can be customized by the user. The selection of default values for \growspace{} were chosen by qualitatively assessing the renderings of plants and selecting the combination of values that created realistic virtual plants. For each \growspace{} parameter we set a range of values. Values ranging from $[0,1]$ are multiplied by the default resolution of the environment of 84 pixels for the control, multi-stageand fairness challenges.
\begin{center}
\begin{table}[h]
\centering
 \begin{tabular}{c c c c} 
 \hline
 \textbf{\growspace{} Parameter} & \textbf{Min} & \textbf{Max} & \textbf{Final}\\ [0.5ex] 
 \hline
 Light Density & 100 & 400 & 200 \\ 
 Initial Light Width & 0.01 & 1 & 0.25 \\
 First Branch Height & 0.05 & 0.5 & 0.2 \\
 Max Branching & 1 & 20 & 8\\
 Branch Length & 0.033 & 0.3 & 0.1 \\
 Branch Thickness & 0.015 & 0.05 & 0.015 \\
 \hline
 \label{tab:sweepgrowspace}
\end{tabular}
\caption{\growspace{} default parameters after performing a sweep}

\end{table}
\end{center}

The hyperparameter sweep for PPO was performed across 3 seeds and values were tested on all 4 challenges. The average episode reward was the metric for determining which combination of hyperparameters allowed for better learning. It is interesting to note that the final hyperparameters are similar to the ALE benchmark for PPO.

\begin{center}
\begin{table}[h]
\centering
 \begin{tabular}{c c c c} 
 \hline
 \textbf{PPO Hyperparameter} & \textbf{Min} & \textbf{Max} & \textbf{Final}\\ [0.5ex] 
 \hline
 Learning rate & 0.00001 & 0.1 & 0.00025 \\ 
 Epsilon & 0.01 & 1 & 0.25 \\
 Gamma & 0.05 & 0.99 & 0.99 \\
 Generalized Advantage Estimation lambda & 0.3 & 0.99 & 0.95\\
  Entropy coefficient & 0.01 & 0.5 & 0.01 \\
 Max grad norm & 0.1 & 0.9 & 0.5 \\
 Number of steps & 1000 & 5000 & 2500 \\
 PPO epoch & 1 & 20 & 4 \\
 Mini batch & 10 & 100 & 32 \\
 Optimizer & adam & sgd & adam\\
 Momentum & 0.95 & 0.99 & 0.95 \\ 
Clip parameter & 0.05 & 0.5 & 0.1\\
 \hline
\end{tabular}
\caption{Range of PPO hyperparameters used while tuning and final values}
\end{table}
\end{center}






\section{Additional Details on Experiments}\label{add_case}
\textbf{Oracles:}
For the control and multi-stage challenges, the target location is known before hand and the light is moved accordingly until the plant has reached the target. For the fairness challenge, the oracle first evaluates if the location of the target is in between, on the right or the left side of the two plants. Once known, the oracle displaces the light to the furthest plant and alternates the lighting to achieve similar growth in both plants towards the target. For the multi-objective challenge, the oracle knows the desired digit to form and displaces the light to fill the digit from bottom to top all while keeping the light within the MNIST digit pixels.

\textbf{Action selection:}
The following are action distributions for the control and multi-stagechallenges for the case studies discussed in Section ~\ref{casestudies}.
\begin{figure}[ht]
\centering
\subfigure[Action distribution easy]{%
    \includegraphics[width=0.4\columnwidth]{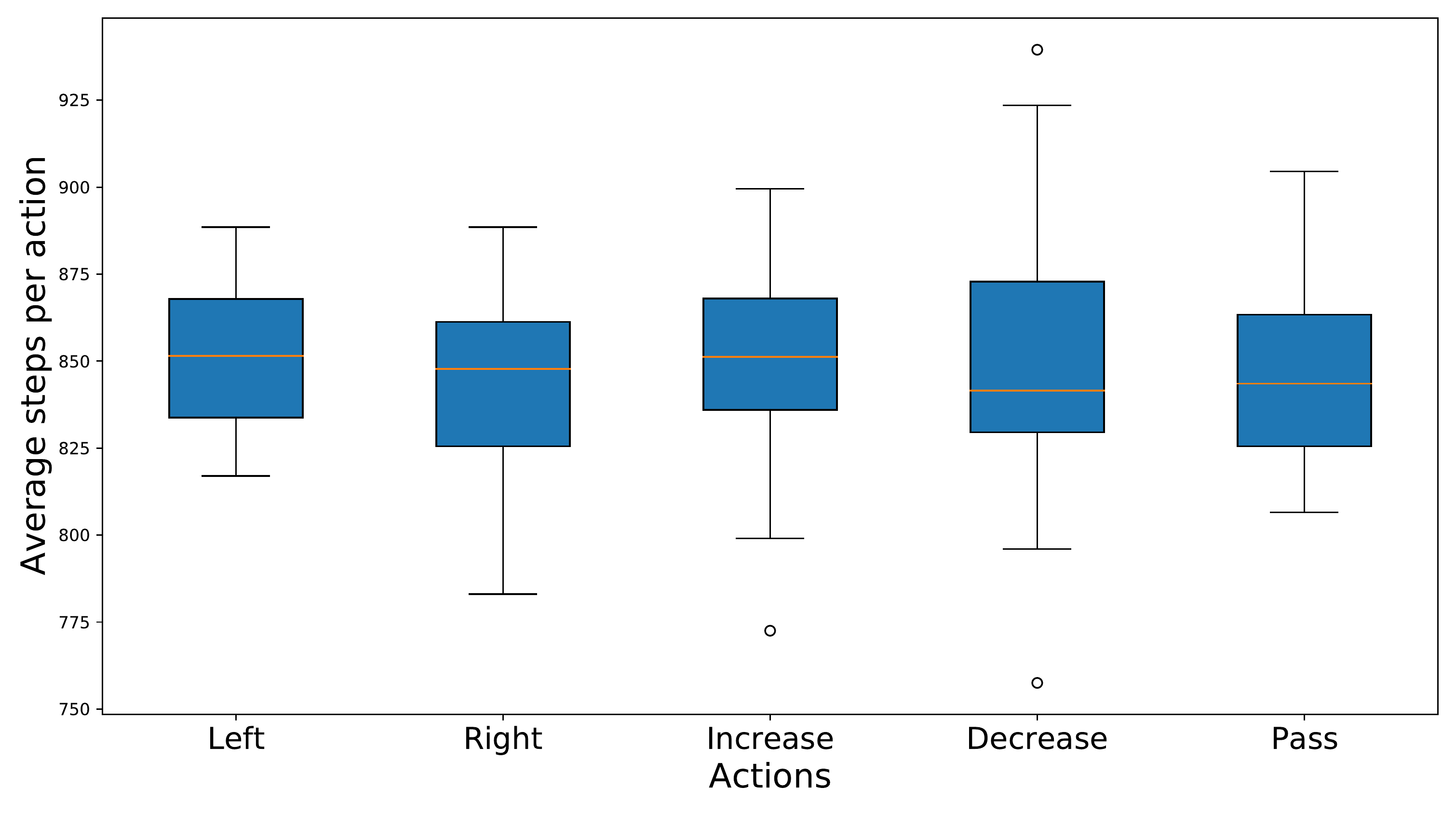}
    \label{Control easy}}
\quad
\subfigure[Action distribution hard]{%
    \includegraphics[width=0.4\columnwidth]{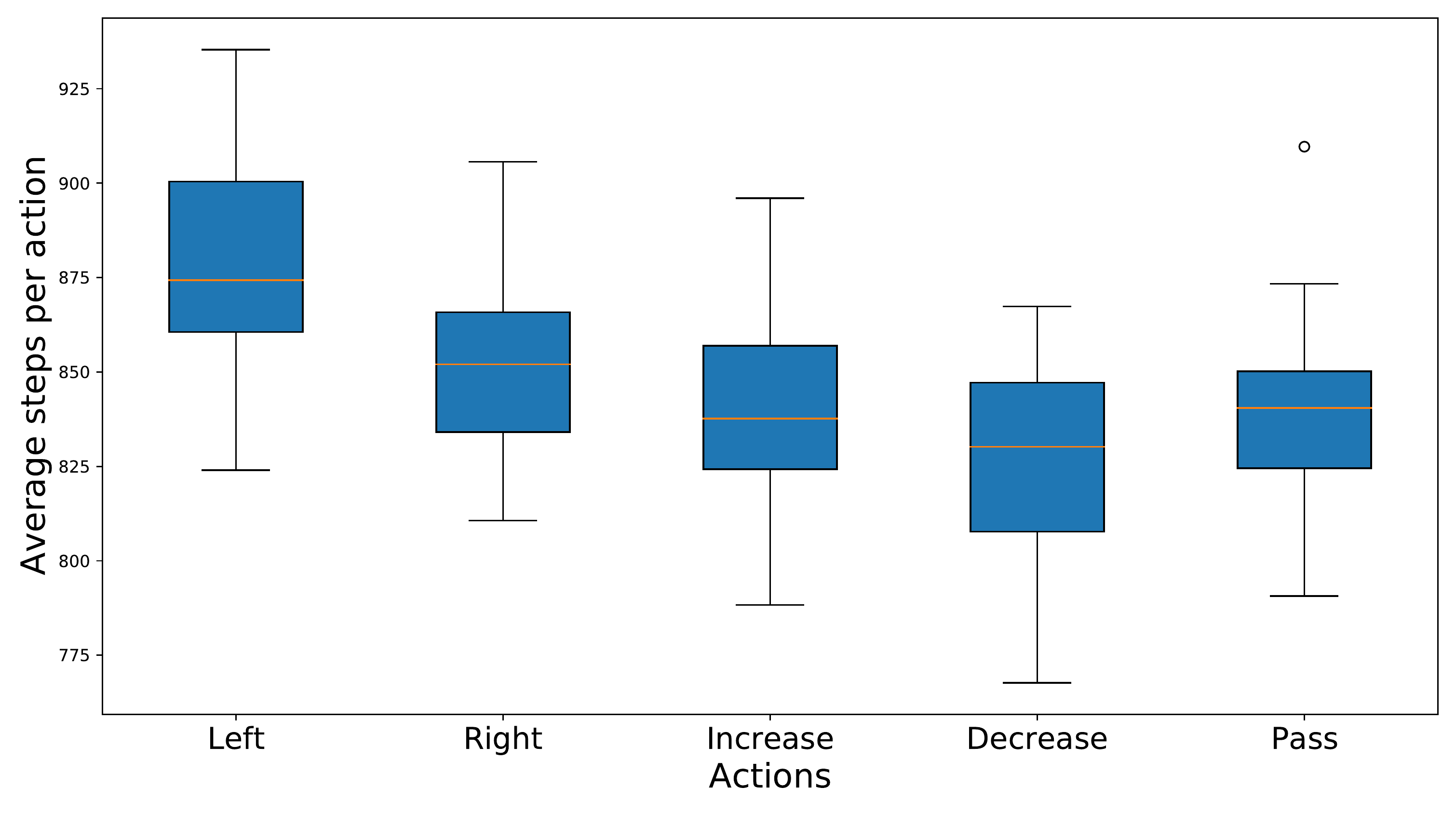}
    \label{Control hard}}
\quad

\caption{Action selection during training for control case studies}
\label{fig:fig_control_case}
\end{figure}

\begin{figure}[ht]
\centering
\subfigure[Action distribution easy]{%
    \includegraphics[width=0.4\columnwidth]{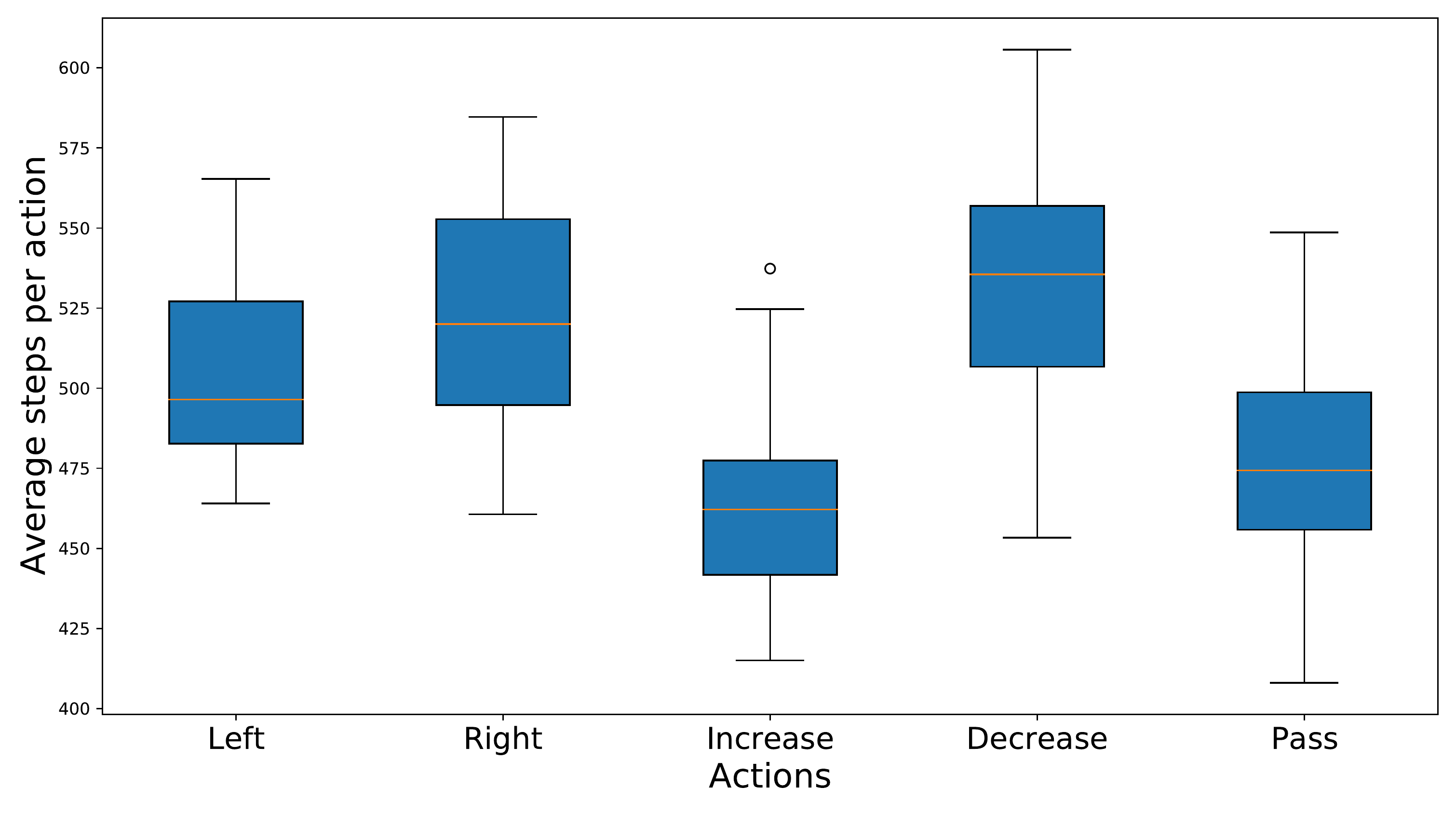}
    \label{Hierarchy easy}}
\quad
\subfigure[Action distribution hard]{%
    \includegraphics[width=0.4\columnwidth]{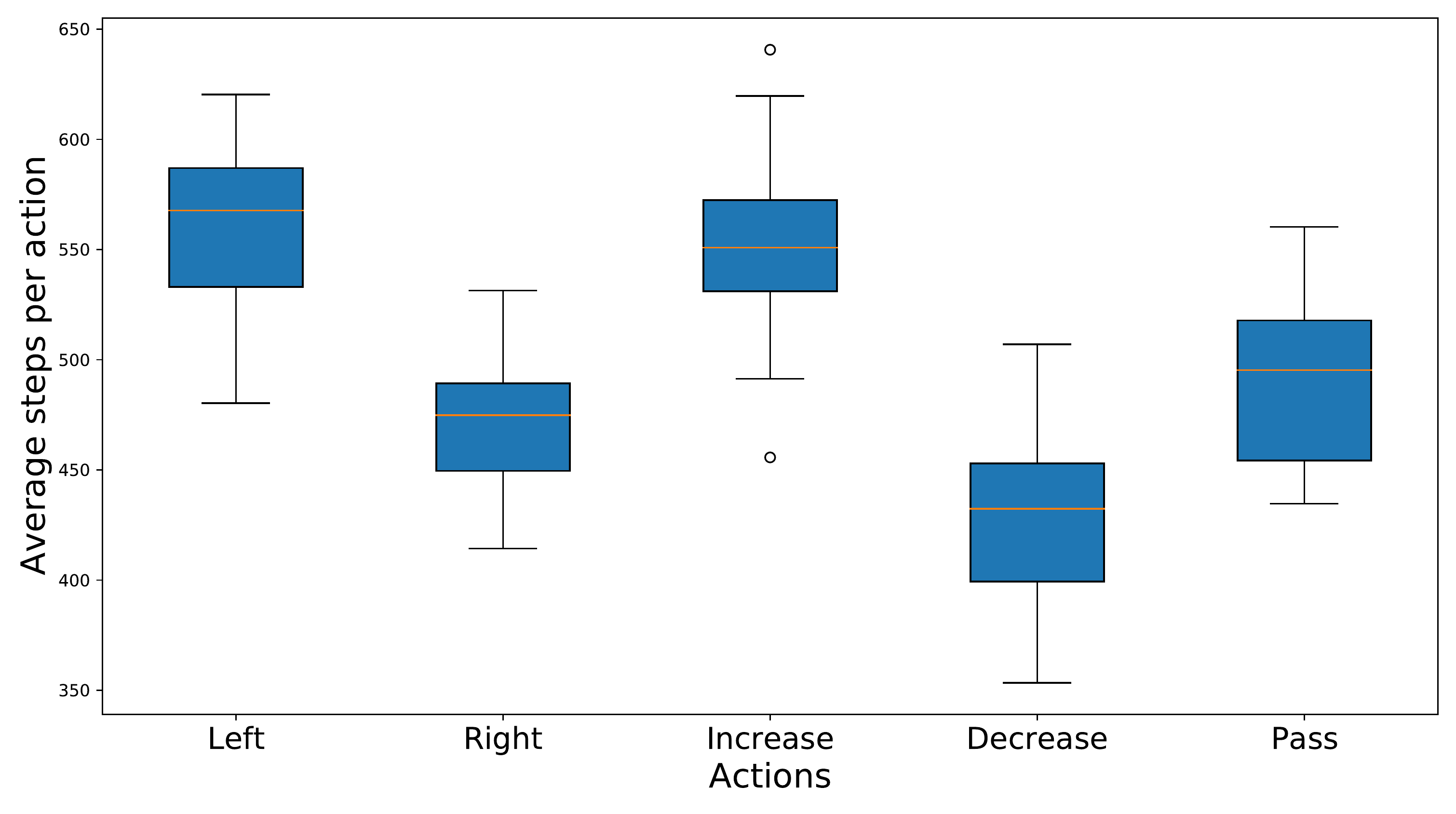}
    \label{Hierarchy hard}}
\quad

\caption{Action selection during training for multi-stagecase studies}
\label{fig:fig_hierarchy_case}
\end{figure}
\begin{figure}[h!]
\centering
\subfigure[Fairness close challenge]{%
    \includegraphics[width=0.4\columnwidth]{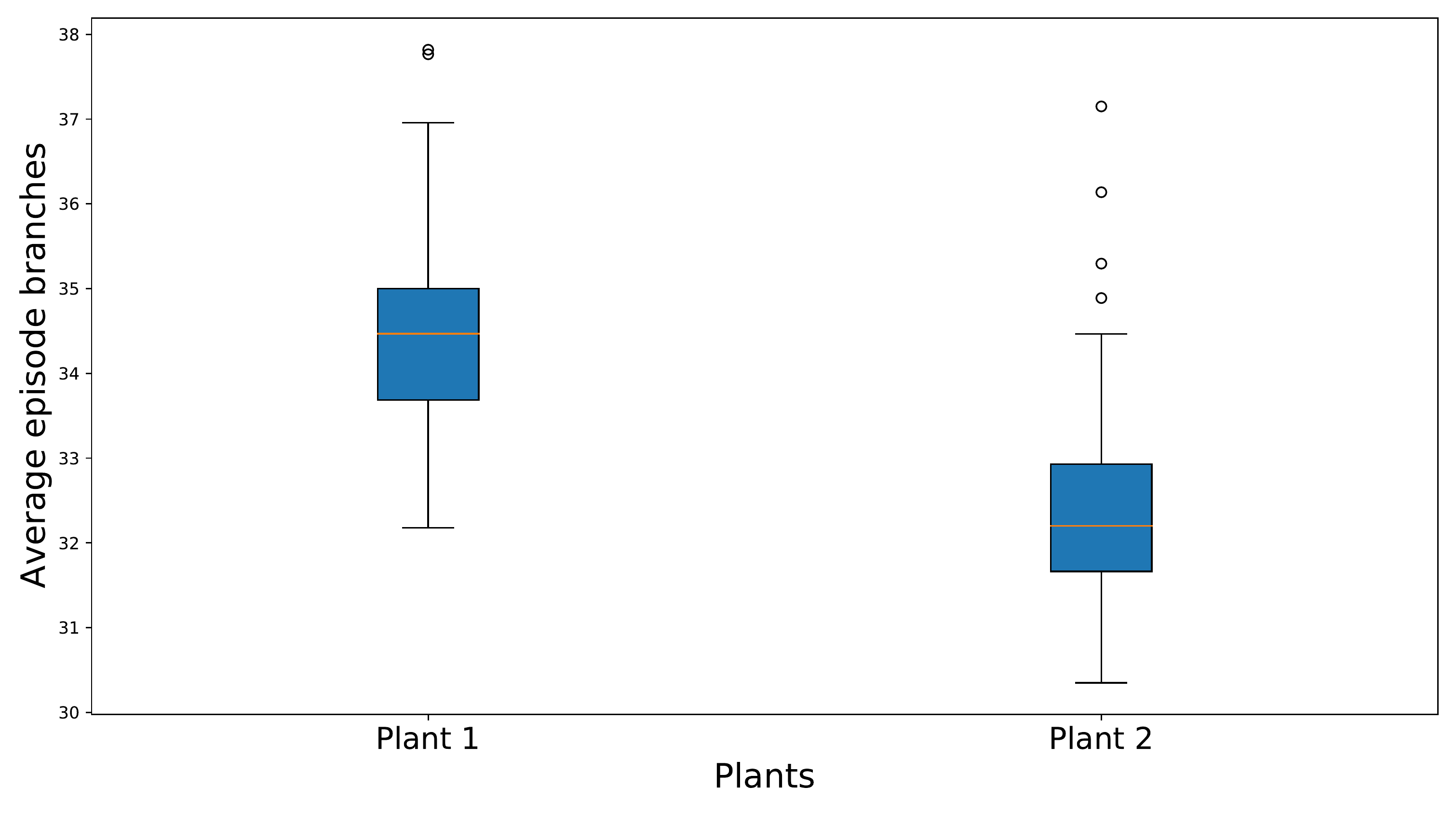}
    \label{faireasy}}
\quad
\subfigure[Fairness above challenge]{%
    \includegraphics[width=0.4\columnwidth]{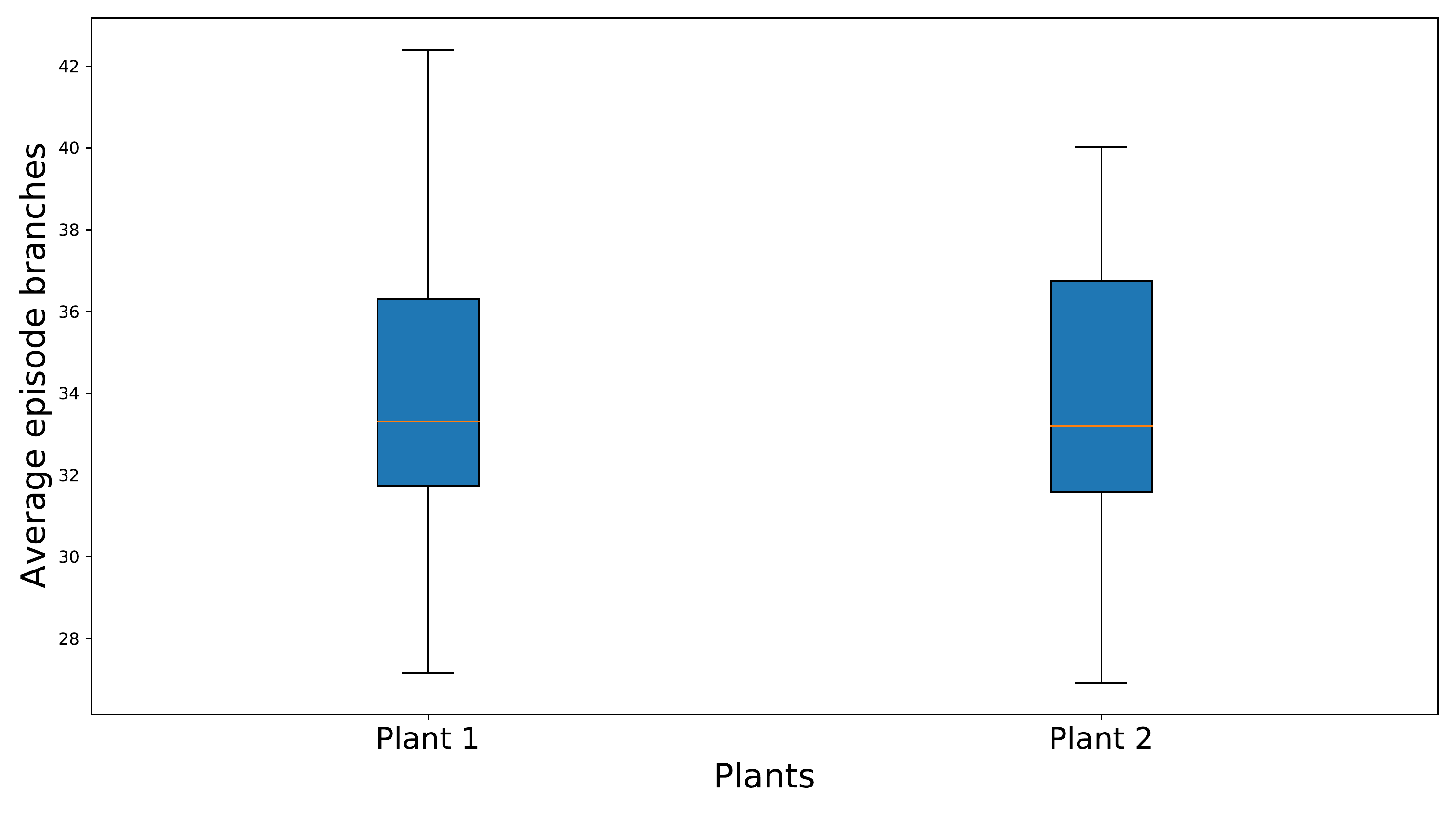}
    \label{fairabove}}
\caption{Easy and hard cases episode branching for fairness challenge}
\label{fig:fairbranches}
\end{figure}

\textbf{Branching:}
In figure ~\ref{fairabove}  we quantitatively validate if the agent is being fair in growing both plants we look into the average number of branches for each plant over the episode. 


\section{Compute Power}\label{compute}
The total amount of compute for this project was of 322 days. This is in part due to game development and the two hyperparameter sweeps done on \growspace{} and PPO. Training was done with 1 GPU and 8 CPU and an average run depending on the challenges in \growspace{} can take from 1 hour and 30 minutes to 3 hours. More details can be found in the reports shared below in Appendix ~\ref{videoagent}. These computing times for one run is comparable to Atari.


\section{Viewing Agent Videos and Experiments}\label{videoagent}
Agent videos and experiments have been tracked with weights and biases \cite{wandb}. Individual dashboards were made for easy access and to better compare case studies within challenges. \begin{itemize}
    \item Control challenge: \url{https://wandb.ai/growspace/control/reports/Control-Challenge--Vmlldzo3NTk1NDk}
    \item Multi-stage challenge: \url{https://wandb.ai/growspace/hierarchy/reports/Hierarchy-Challenge--Vmlldzo3NTk1MzI}
    \item Fairness challenge: \url{https://wandb.ai/growspace/fairness/reports/Project-Dashboard--Vmlldzo3NTk1NjI}
    \item Multi-objective challenge: \url{https://wandb.ai/growspace/mnistexperiments/reports/Project-Dashboard--Vmlldzo3NTk1NTk}
\end{itemize}

\section{Changes from first round to second round for submission}
In this section we provide clarification on changes we have addressed since our first submission.We address the following:
\begin{itemize}
    \item  Justification on our selection of SCA and demonstrate that it is not artificial and that indeed it can provide realistic renderings (Figure ~\ref{overviewfig}), and that it can encompass a real-world phenomena of phototropic behaviour in plants (Section ~\ref{plant branching} and Appendix ~\ref{add_grow}).
    \item We equally have changed our Figure ~\ref{overviewfig} to better demonstrate the use of SCA and its applications outside of \growspace{}
    \item In the introduction we highlight why benchmarks with above human-like performances no longer present challenges to the RL community.
    \item In Section ~\ref{experimentsresults} we have included the figure demonstrating the overall performance of 3 baselines on \growspace{} control setting, we present A2C, PPO and Rainbow DQN (see Figure ~\ref{fig:control-baselines}. This figure was previously in 
    \item In Section ~\ref{experimentsresults} we clarify the case study on the multi-objective challenge and provide greater depth on the difficulty of this setting. 
    \item A new figure illustrating the episode initialization for the case studies presented has been made, which includes plots from previous figures.

\end{itemize}
\end{document}